\documentclass[letterpaper, 10 pt, conference]{ieeeconf}  

\IEEEoverridecommandlockouts                              

\usepackage{graphics} 
\usepackage{epsfig} 
\usepackage{mathptmx} 
\usepackage{times} 
\usepackage{amsmath} 
\usepackage{amssymb}  

\usepackage[hidelinks]{hyperref}
\usepackage{subcaption}

\title{\LARGE \bf
MIMo grows! Simulating body and sensory development \\ in a multimodal infant model
}

\author{
Francisco M. López*\(^{1}\),
Miles Lenz*\(^{2}\),
Marco G. Fedozzi*\(^{3}\),
Arthur Aubret\(^{4}\),
and Jochen Triesch\(^{5}\)
\thanks{
* Equal contributions. This research was supported by ``The Adaptive Mind'' funded by the Excellence Program of the Hessian Ministry of Higher Education, Research, Science and the Arts, Germany, and by the Deutsche Forschungsgemeinschaft (DFG project “Abstract REpresentations in Neural Architectures (ARENA)”). JT was supported by the Johanna Quandt foundation.
}
\thanks{
\(^1\) Francisco M. López is with the Frankfurt Institute for Advanced Studies, Germany, and with the Xidian-FIAS International Joint Research Center, Germany. \footnotesize{\texttt{lopez@fias.uni-frankfurt.de}}
}%
\thanks{
\(^2\) Miles Lenz is with the Goethe University Frankfurt, Germany.  \footnotesize{\texttt{lenz@fias.uni-frankfurt.de}}
}%
\thanks{
\(^3\) Marco G. Fedozzi is with the University of Genova, Italy, and with Instituto Italiano di Tecnologia, Italy. \footnotesize{\texttt{marco.fedozzi@iit.it}}
}%
\thanks{
\(^4\) Arthur Aubret is with the Frankfurt Institute for Advanced Studies, Germany, and with the Xidian-FIAS International Joint Research Center, Germany. \footnotesize{\texttt{aubret@fias.uni-frankfurt.de}}
}%
\thanks{
\(^5\) Jochen Triesch is with the Frankfurt Institute for Advanced Studies, Germany, and with the Goethe University Frankfurt, Germany. \footnotesize{\texttt{triesch@fias.uni-frankfurt.de}}
}%
}

\begin{document}

\maketitle
\thispagestyle{empty}
\pagestyle{empty}

\begin{abstract}

Infancy is characterized by rapid body growth and an explosive change of sensory and motor abilities. However, developmental robots and simulation platforms are typically designed in the image of a specific age, which limits their ability to capture the changing abilities and constraints of developing infants. To address this issue, we present MIMo v2, a new version of the multimodal infant model. It includes a growing body with increasing actuation strength covering the age range from birth to 24 months. It also features foveated vision with developing visual acuity as well as sensorimotor delays modeling finite signal transmission speeds to and from an infant's brain. Further enhancements of this MIMo version include an inverse kinematics module, a random environment generator and updated compatiblity with third-party simulation and learning libraries. Overall, this new MIMo version permits increased realism when modeling various aspects of sensorimotor development. The code is available on the official repository (\href{https://github.com/trieschlab/MIMo}{https://github.com/trieschlab/MIMo}).
\end{abstract}

\section{INTRODUCTION}

Piaget described the first two years of a child's life as the sensorimotor stage of cognitive development \cite{winstanley2023stages}. During this time, infants learn about themselves and the world around them. They acquire this knowledge through interactions, such that they can build internal representations of their environments on the basis of active exploration and curiosity. This period is characterized by rapid body growth \cite{anthrokids} and an explosive change of their sensory and motor abilities \cite{corbetta2021perception}. As their bodies change, so do the experiences they have access to: they coordinate their sensory systems \cite{tondel2008accommodation}, reach for and manipulate objects \cite{corbetta2018learning}, acquire the ability to sit \cite{harbourne2013sit}, and eventually locomote to discover new interesting places \cite{adolph2017development}. In sum, the gulf that separates a newborn and a toddler poses a conundrum for developmental scientists: is it really possible to understand early cognitive development by treating infancy as a single stage? The answer from the perspective of developmental psychology and neuroscience is quite clear: no \cite{corbetta2021perception,winstanley2023stages}. And yet, when it comes to developmental robotics and artificial intelligence, most works aim to describe behavioral phenomena that take place at specific ages within the first two years of life using a single robotic or virtual artificial agent that shares only partially the capabilities and constraints of a real infant at that age. Examples of such agents include the iCub \cite{metta2010icub}, NICO \cite{kerzel2017nico}, MIMo \cite{mattern2024mimo}, and the fetus simulator \cite{kim2022simulating}.

\begin{figure}[!b]
    \centering
    \includegraphics[width=1\linewidth]{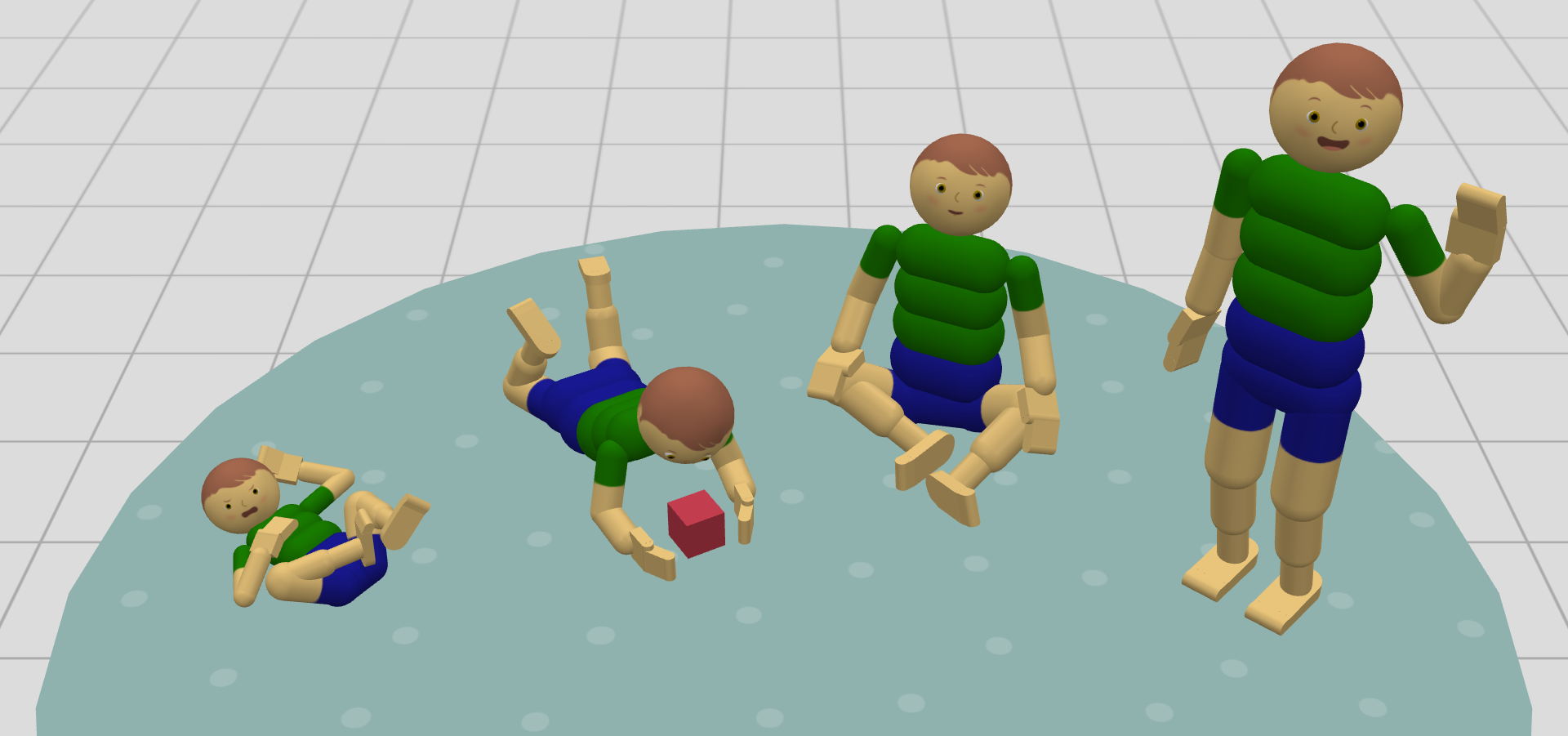}
    \caption{Illustration of MIMo at different ages. From left to
right: 0, 6, 18, and 24 months.}
    \label{fig:mimos}
\end{figure}

\begin{figure*}[!t]
    \centering
    \begin{subfigure}[t]{0.31\textwidth}
        \centering
        \includegraphics[width=\linewidth]{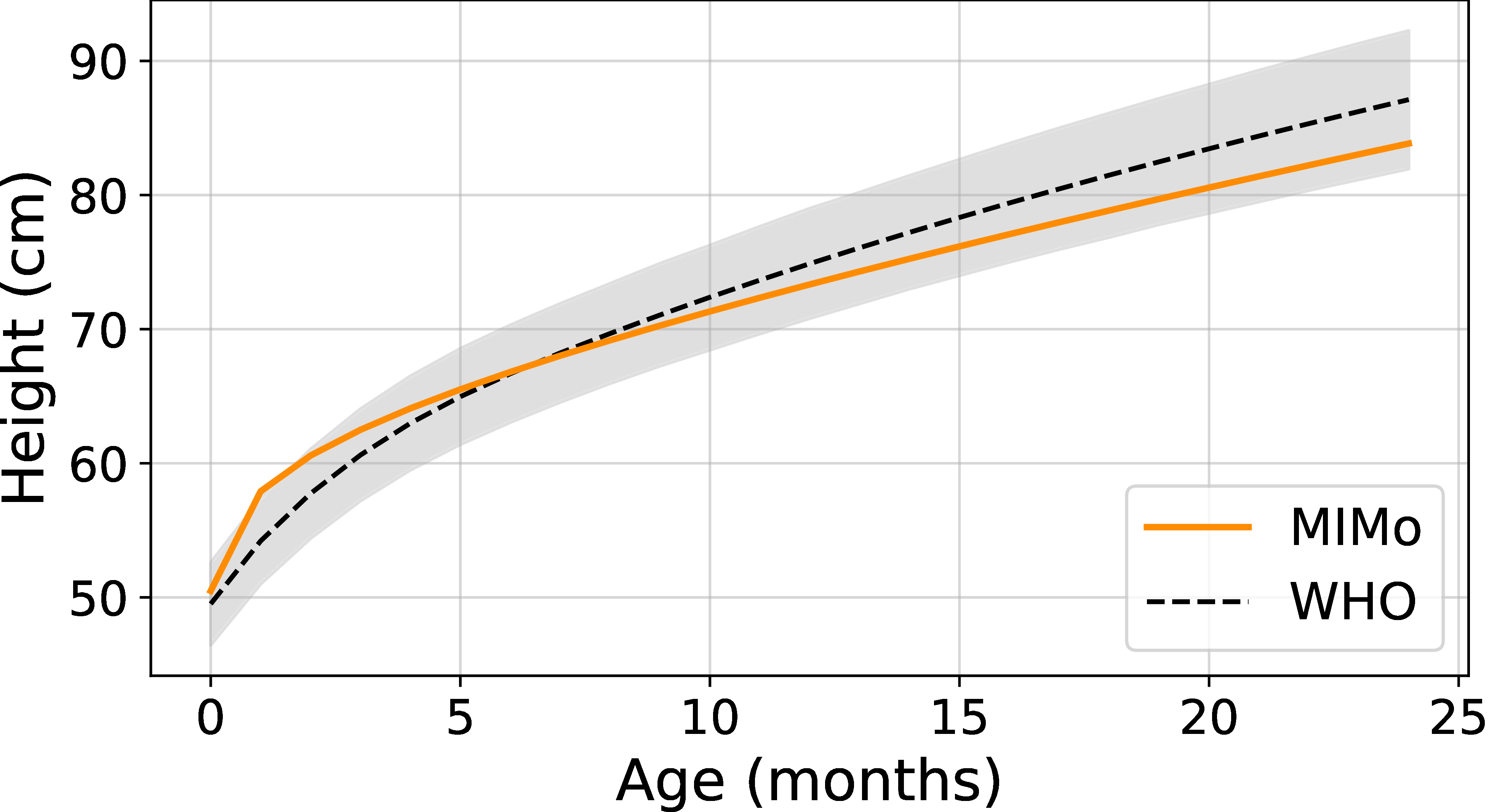}
        \caption{MIMo's body height}
    \end{subfigure}
    \hfill
    \begin{subfigure}[t]{0.31\textwidth}
        \centering
        \includegraphics[width=\linewidth]{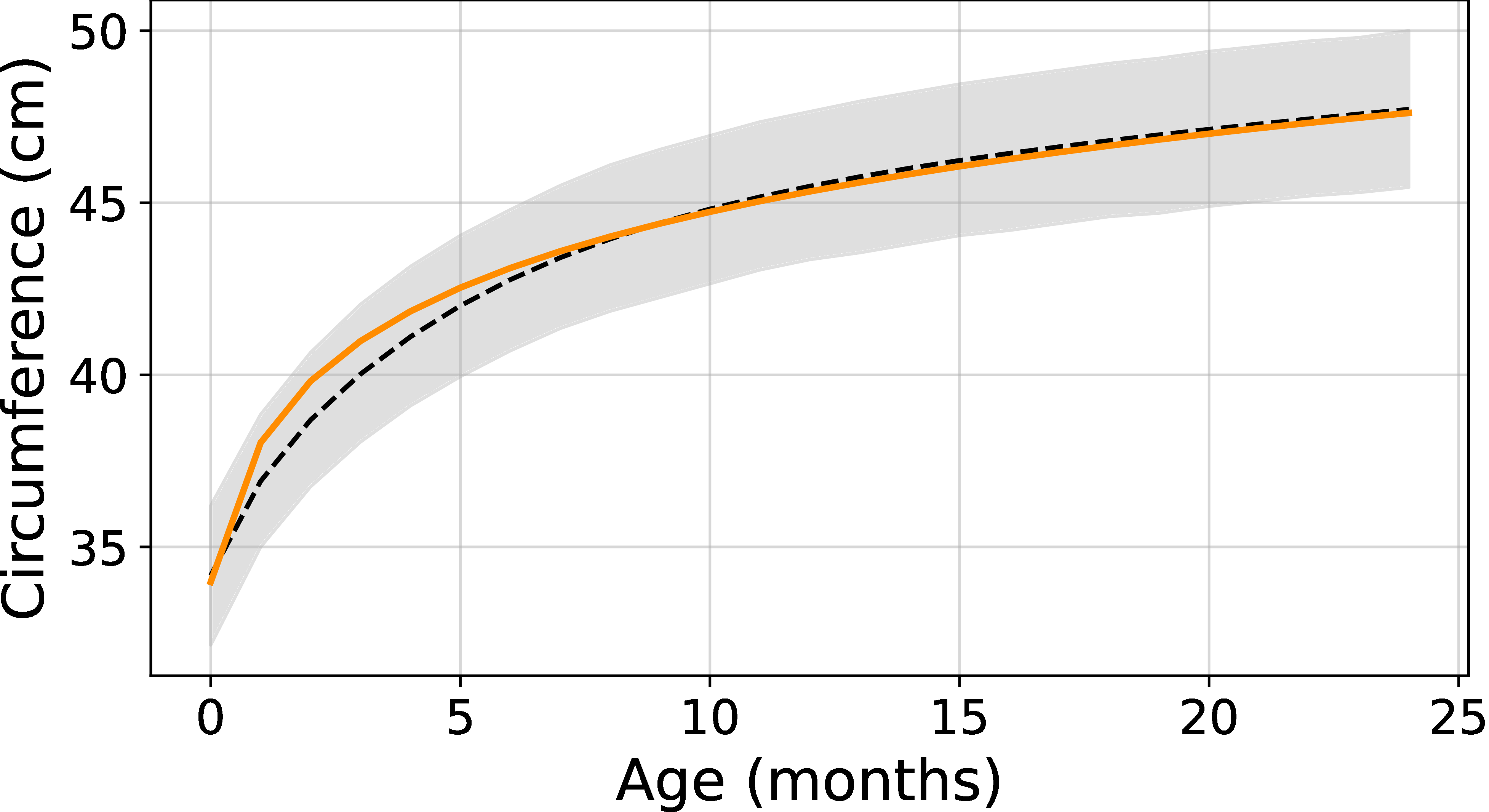}
        \caption{MIMo's head circumference}
    \end{subfigure}
    \hfill
    \begin{subfigure}[t]{0.31\textwidth}
        \centering
        \includegraphics[width=\linewidth]{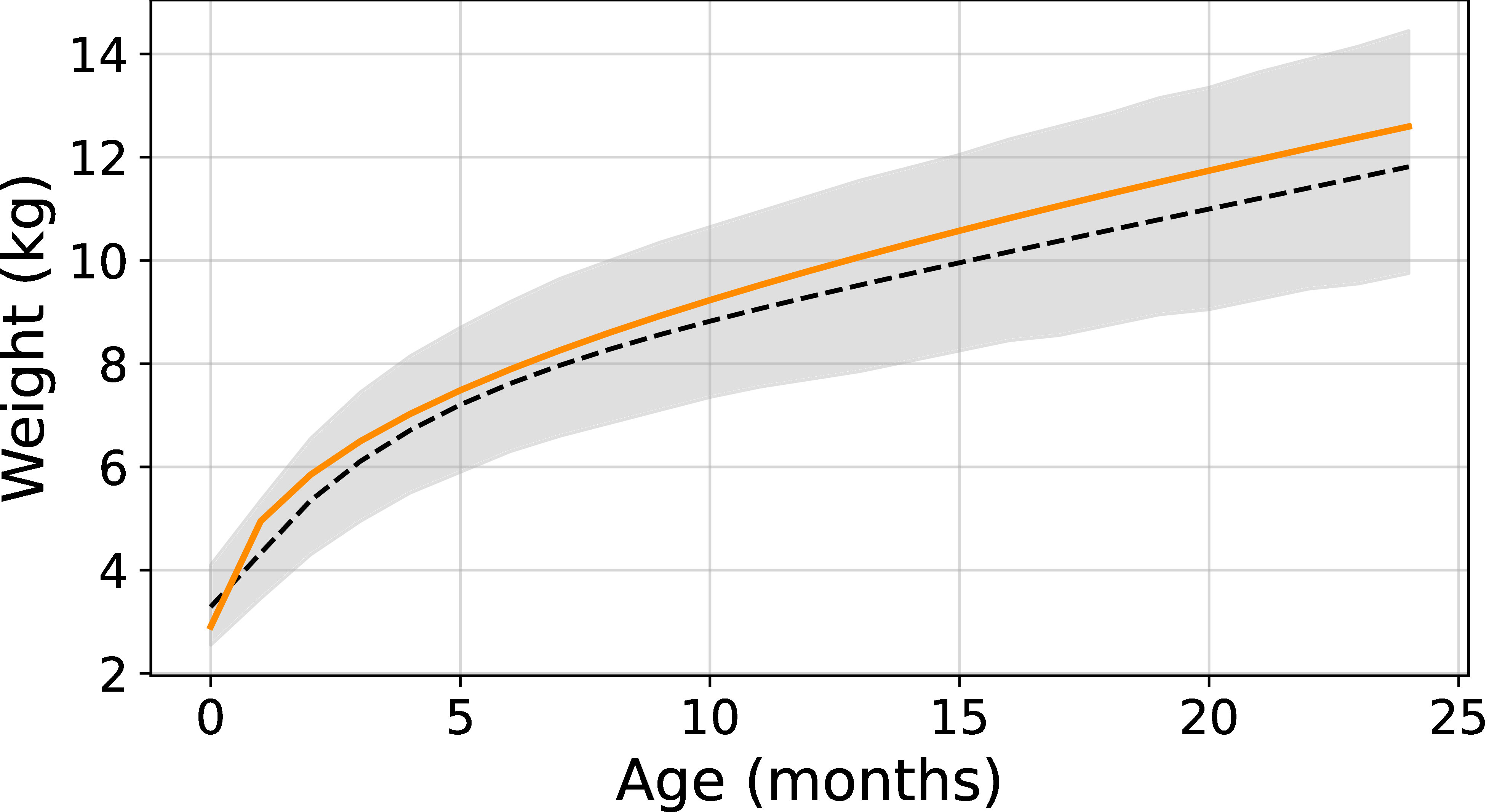}
        \caption{MIMo's body mass}
    \end{subfigure}
    \\[1em]
    \begin{subfigure}[t]{0.48\textwidth}
        \centering
        \includegraphics[width=\linewidth]{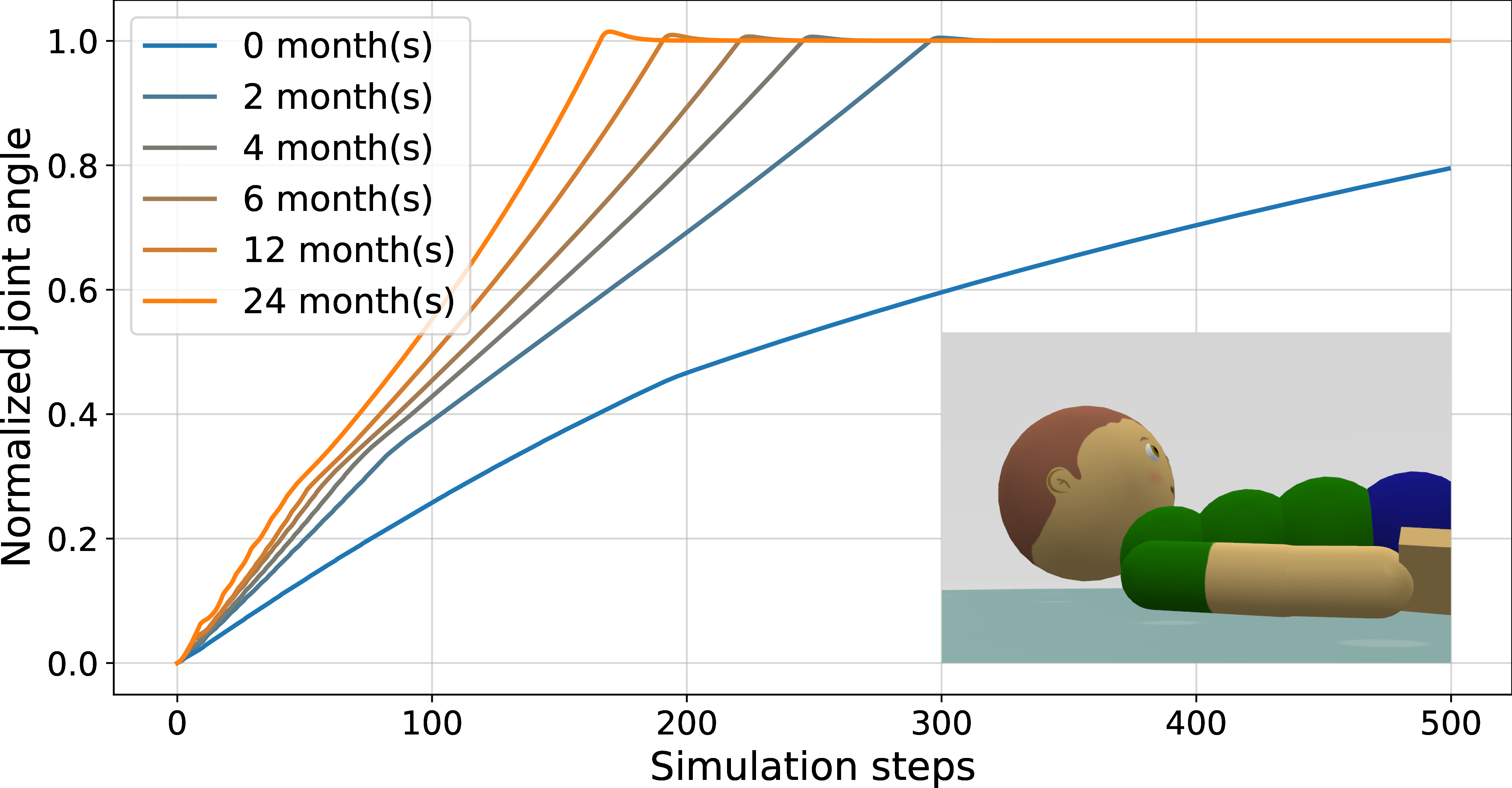}
        \caption{Head lifts}
    \end{subfigure}
    \hfill
    \begin{subfigure}[t]{0.48\textwidth}
        \centering
        \includegraphics[width=\linewidth]{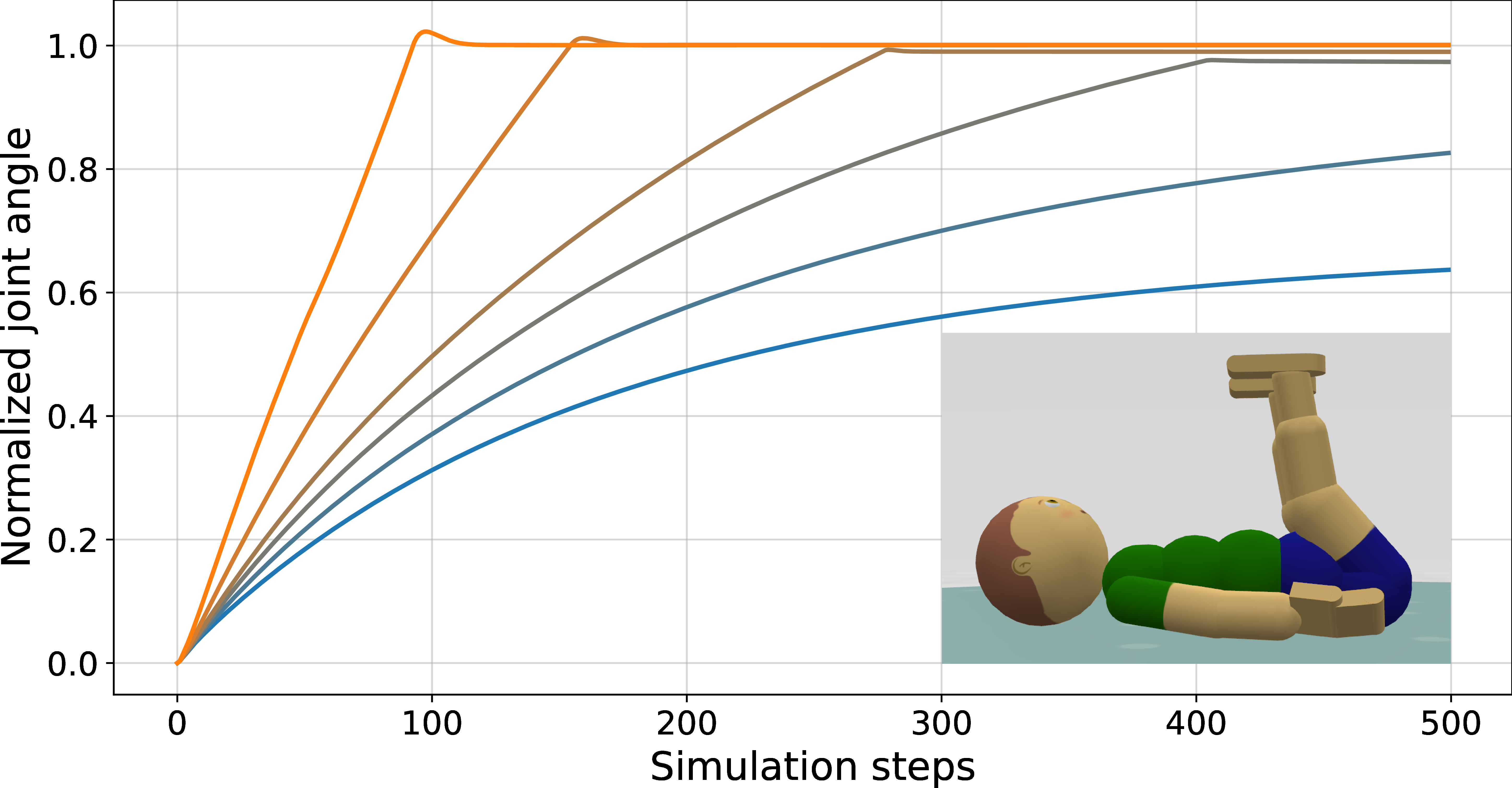}
        \caption{Leg lifts}
    \end{subfigure}
    \caption{MIMo's growth results. Top: MIMo's developing body compared with real infant data reported by the WHO \cite{worldhealthorganization2006WHO, worldhealthorganization2007WHO} (mean and 5th-95th percentiles range shown). Bottom: strength tests with MIMo lifting his head (left) and legs (right) from a supine position. Insets show a 6-month-old MIMo performing both behaviors.}
    \label{fig:growth}
\end{figure*}

In this work, we address this gap with MIMo v2, an extension of the multimodal infant model \cite{mattern2024mimo}. MIMo is a simulation platform for studying early cognitive development, originally designed after an 18-month-old infant and successfully deployed in several research projects to model visual control \cite{lopez2024self, raabe2023saccade}, body movements \cite{fiala2023retargeting}, and eye-hand coordination \cite{lopez2023eye}. The extensions included in this update, which are described in further detail in the remainder of the paper, are the following: a growing body covering the age range from birth to 24 months with corresponding mass and actuation strength; a more realistic visual perception with foveation and age-dependent visual acuity; and sensorimotor delays to capture finite signal transmission speeds along nerve fibers to and from an infant's brain. Additionally, MIMo v2 includes an inverse dynamics controller and a random environment generator. Overall, this MIMo update permits increased realism when modeling the sensorimotor and cognitive development of infants {\em in silico}.

\section{Overview of MIMo}

MIMo \cite{mattern2024mimo} is a multimodal infant model built on top of the MuJoCo platform \cite{todorov2012mujoco} that can be run using Gymnasium environments \cite{towers2024gymnasium} for faster than real-time simulations of infant behaviors. His original body is composed of simple geometrical primitives adjusted to match the dimensions of an average 18-month-old child taken from the AnthroKids measurements \cite{anthrokids}. Two different versions of MIMo are available: one with mitten-like hands, that has 44 degrees of freedom,  and a computationally more expensive one with five-fingered hands and a total of 88 degrees of freedom. 

MIMo has access to four sensory modalities: proprioception, vestibular system, vision, and touch. Proprioceptive observations contain the position, velocity, force and limit values for each joint in MIMo's body. MIMo has access to binocular vision from cameras located in his eyes.
Touch perception is implemented as a full-body virtual skin with uniformly distributed MuJoCo haptic sensors with densities defined per body part, allowing for higher density in the hands and fingers. These sensors register contacts with external objects or other parts of MIMo's body.

MIMo can move using three different types of actuation models: positional controllers that directly select the angle of each joint, torque controllers where actions are modeled as a motor with a spring-damper system, and muscle controllers where joints are actuated by differentially activating antagonistic muscle pairs. 

For a detailed description of MIMo's design, multimodal perception, and actuation models, please see \cite{mattern2024mimo, mattern2022mimo}. 

\section{Body Growth}

MIMo v2 includes a new growing body that aims to capture the changes in physical size, strength, and constraints endured by an infant during their first two years of life. This is achieved by updating MIMo's body size, body mass, and actuation models, using data from real infants reported in the literature, to guarantee a continuous growth from 0 to 24 months. MIMo's age can now be set at the beginning of an experiment, and the corresponding embodiment will be rendered and simulated.

\begin{figure*}[!t]
    \centering
    \begin{subfigure}[t]{0.21\textwidth}
        \centering
        \includegraphics[width=\linewidth]{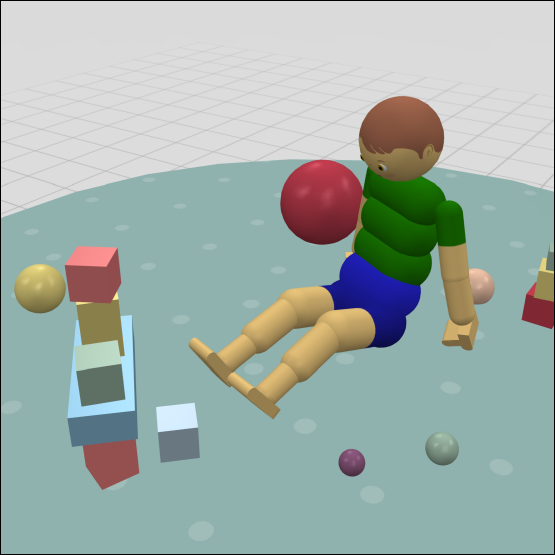}
        \caption{External view}
    \end{subfigure}
    \hfill
    \begin{subfigure}[t]{0.21\textwidth}
        \centering
        \includegraphics[width=\linewidth]{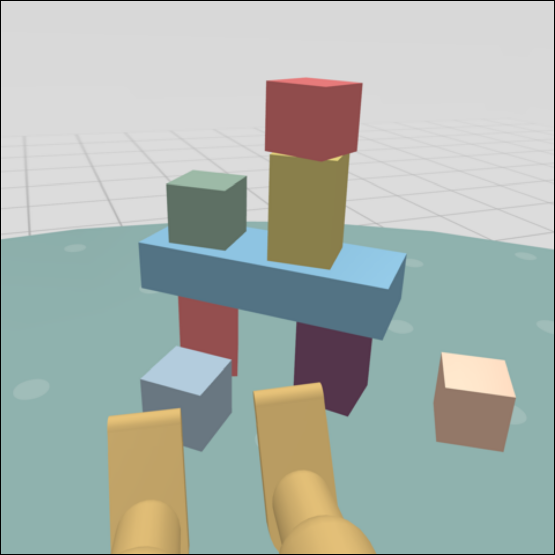}
        \caption{MIMo's original vision}
    \end{subfigure}
    \hfill
    \begin{subfigure}[t]{0.21\textwidth}
        \centering
        \includegraphics[width=\linewidth]{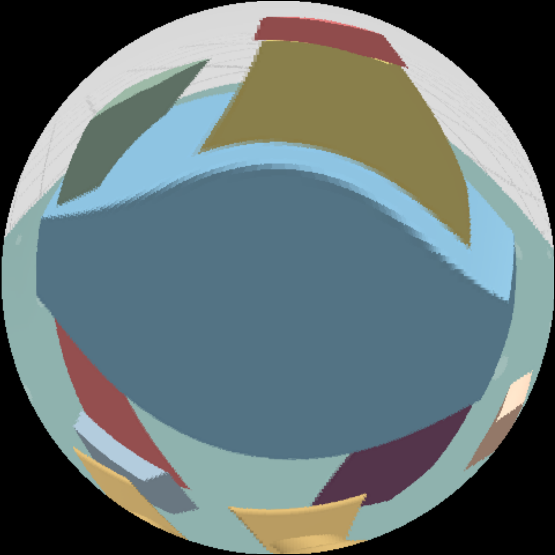}
        \caption{MIMo's foveated vision}
    \end{subfigure}
    \hfill
    \begin{subfigure}[t]{0.31\textwidth}
        \centering
        \includegraphics[width=\linewidth]{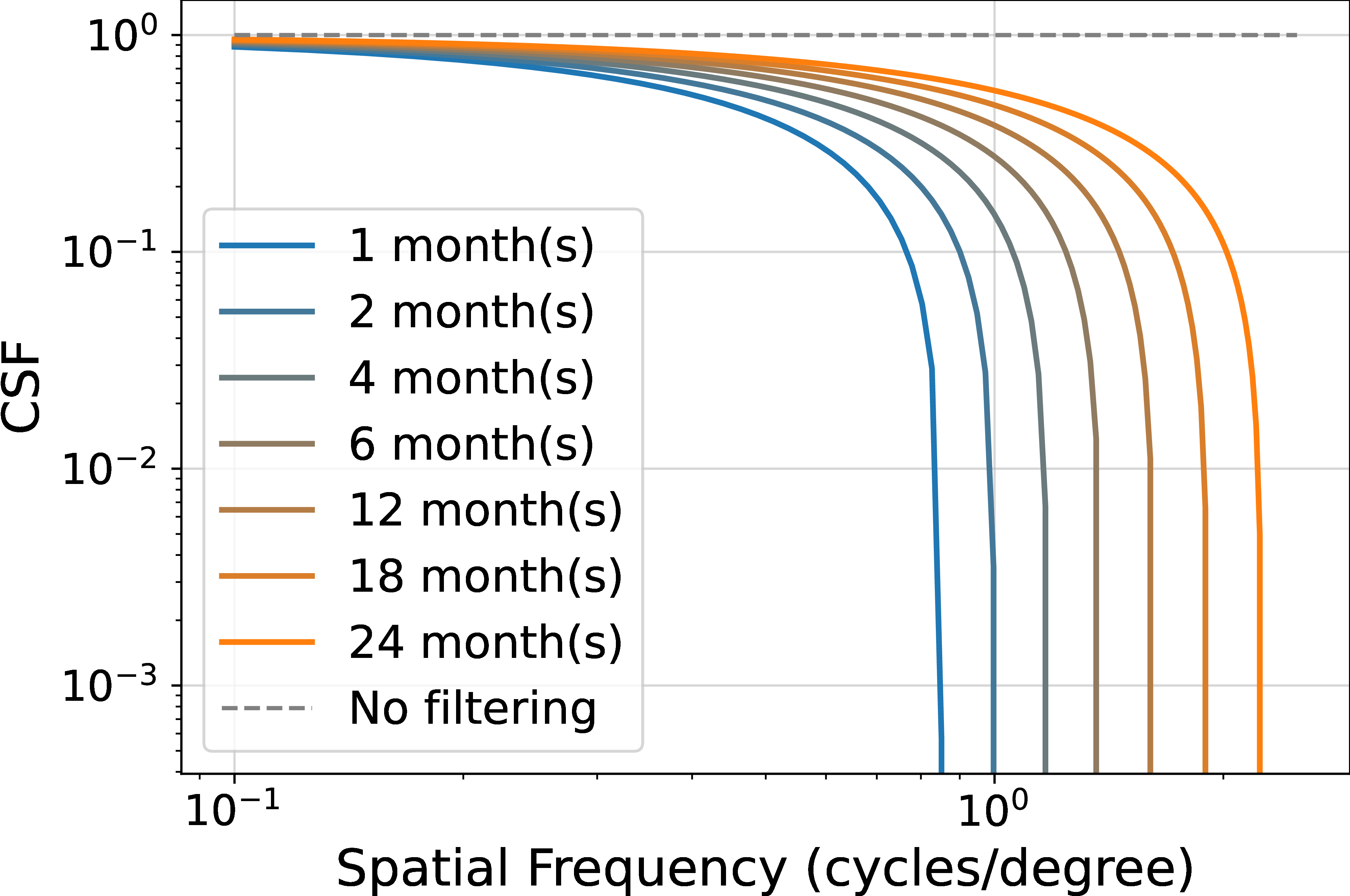}
        \caption{CSF approximations}
    \end{subfigure}
    \\[1em]
    \begin{subfigure}[t]{0.18\textwidth}
        \centering
        \includegraphics[width=\linewidth]{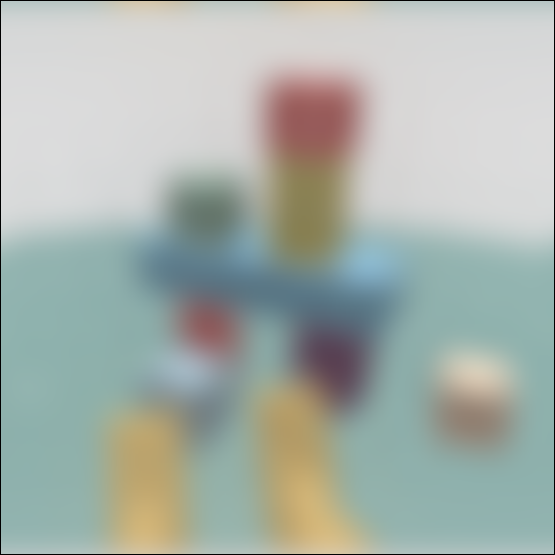}
        \caption{1 month}
    \end{subfigure}
    \hfill
    \begin{subfigure}[t]{0.18\textwidth}
        \centering
        \includegraphics[width=\linewidth]{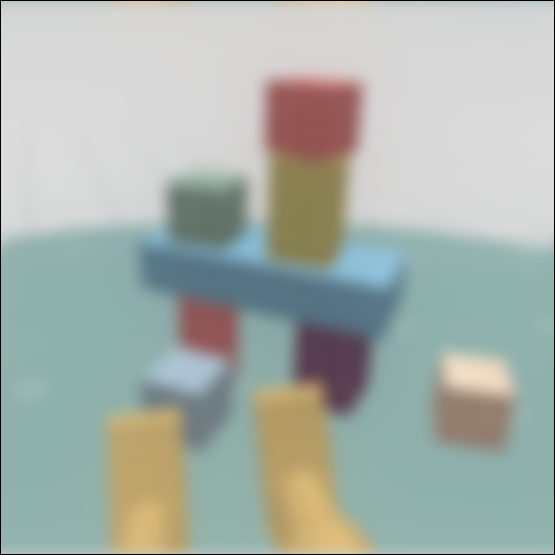}
        \caption{2 months}
    \end{subfigure}
    \hfill
    \begin{subfigure}[t]{0.18\textwidth}
        \centering
        \includegraphics[width=\linewidth]{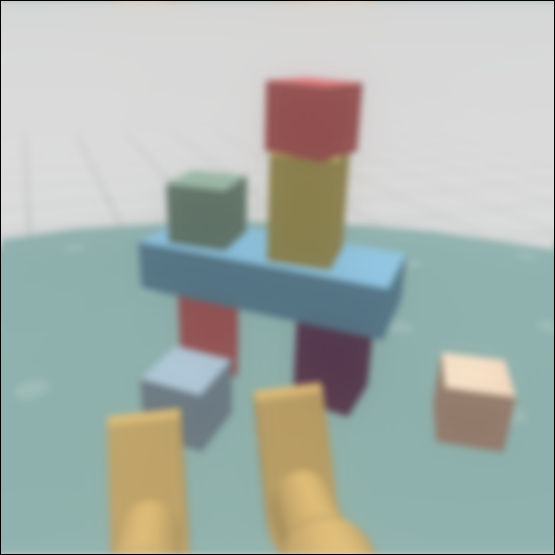}
        \caption{4 months}
    \end{subfigure}
    \hfill
    \begin{subfigure}[t]{0.18\textwidth}
        \centering
        \includegraphics[width=\linewidth]{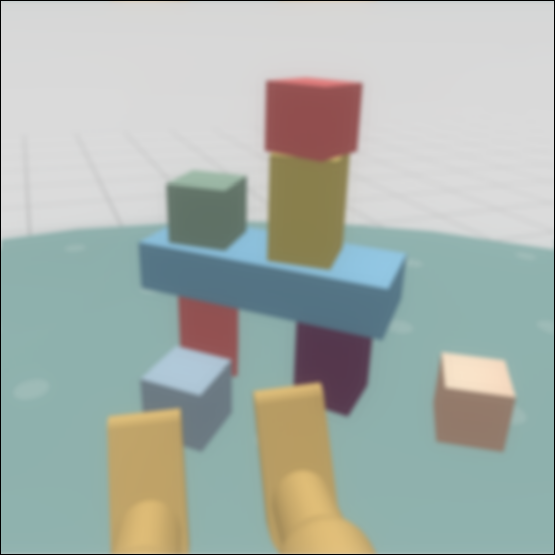}
        \caption{6 months}
    \end{subfigure}
    \hfill
    \begin{subfigure}[t]{0.18\textwidth}
        \centering
        \includegraphics[width=\linewidth]{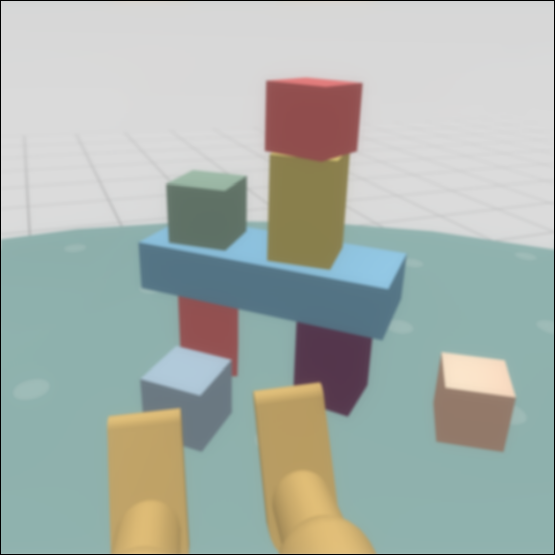}
        \caption{12 months}
    \end{subfigure}
    \\
    \caption{MIMo's new vision module with visual acuity filtering and foveation, here rendered with a field-of-view of 60º.}
    \label{fig:acuity}
\end{figure*}

\subsection{Body size}

Since MIMo's body is made up of simple geometrical primitives, each of these needs to be adjusted to match the desired body size. Infant measurements were taken from the AnthroKids measurements \cite{anthrokids}. We fit the average value of each measurement of a body part to a logarithmic function of the age in months for interpolation:
\begin{equation}
    f(x) = a \ \log(x + b) + c,
\end{equation}

\noindent where the values of \(a\), \(b\), and \(c\) are fit to match the experimental data, with the additional constraint \(b\geq0.1\) to avoid values too close to zero inside the logarithm. An example of the outcome function is shown in Fig.~\ref{fig:growth}(b) for MIMo's head circumference. Values not reported in the AnthroKids measurements, e.g. the height of the feet and hands, were inferred from the relative changes of the available measurements. Besides changing the dimensions of the geometries, their positions within MIMo's body needed to be adjusted to match the necessary joints. Once all relevant parameters are set, a function modifies them in a temporary copy of MIMo's default XML file that is read by MuJoCo when generating the environment. 

Additionally, individual body parts can be modified to fit specific measurements after setting MIMo to a given age, namely in order to recreate a particular infant's body. This can be of significance, for example, when using MIMo to retarget behaviors measured through motion capture of real children \cite{fiala2023retargeting}. 

As a validation for MIMo's body growth, we compare MIMo's age-dependent height and head circumference with data reported by the World Health Organization (WHO) \cite{worldhealthorganization2006WHO}. The results shown in Fig.~\ref{fig:growth}(a)-(b) confirms that MIMo's development falls within one standard deviation of the development of infants worldwide. 

\subsection{Body mass}

While there are many reports of normally-developing infants' weights (e.g. \cite{worldhealthorganization2007WHO}), estimating the mass of each body part individually is a difficult task. Here, we opt for a simple solution: assuming that the density of each body part remains constant during development, the volume of the primitive geometries is indicative of their corresponding mass. Thus, we use the densities of the geometries provided by the original MIMo as a reference and scale the mass according to the relative volume change. As a result, MIMo's weight increases with age similarly to the infants measured in the WHO report \cite{worldhealthorganization2007WHO} (see Fig.~\ref{fig:growth}(c)).

\subsection{Actuation models}

As infants grow, so do their muscles. However, experimental measurements of infants' strengths are rare and mostly focus on their ability to accomplish tasks that involve the activation of multiple muscles \cite{eliks2022alberta}. MIMo's original actuation models were parametrized with the few results available in the literature for infants of 18 months of age, and the remaining parameters were extrapolated from studies for older children. In this work, we opt for an alternative approach, based on the correlation between muscle size and muscle strength \cite{jaric2003role}. MIMo's strength, as indicated by the gear value of the joints, is scaled with the volume of the actuating body part. The body part's volume, but not the more conventional cross-sectional area, is well defined for all joints.  Hence, as MIMo's body develops, each body part increases in size and the strength of the actuators that depend on that body part increase proportionally. This change affects both the torque and the muscle controllers equally, while the positional controllers have no inherent strength. 

As a proof of concept experiment of the updated actuated models, we evaluate MIMo's ability to perform two common behaviors of young infants: lifting the head, typically achieved at the age of 4 months \cite{van2021active}, and lifting a leg, which can happen spontaneously at young ages but normally develops around the age of 3 months \cite{angulo2002three}. MIMo is positioned in a supine position and the corresponding torque actuators (head and right leg) are activated at maximum capacity. The experiment is repeated at multiple ages and the joint angles over 500 timesteps (each corresponding to 5~ms) are recorded. The results shown in Fig.~\ref{fig:growth}(d)-(e) reveal that as MIMo grows, the speed at which he reaches the maximum value of the joint decreases, i.e. he gets stronger. Both behaviors are achieved earlier than in real infants, although such comparisons are not particularly rigorous since, unlike the infants, in this case MIMo does not need to learn this motor control.

\begin{figure*}[!t]
    \centering
    \begin{subfigure}[t]{0.21\textwidth}
        \centering
        \includegraphics[width=\linewidth]{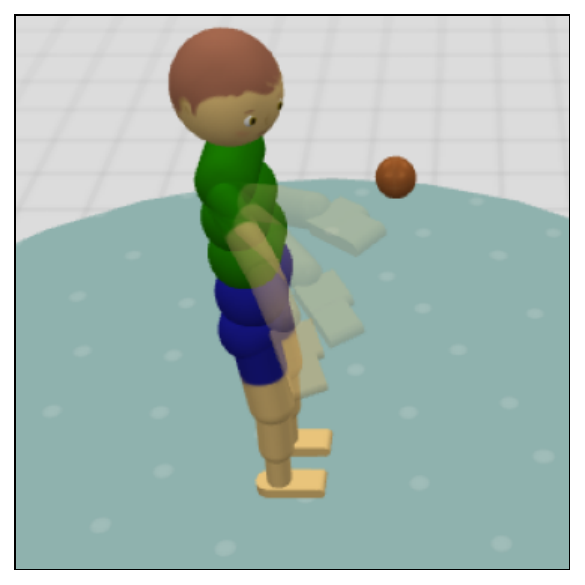}
        \caption{MIMo reaching the target}
    \end{subfigure}
    \hfill
    \begin{subfigure}[t]{0.275\textwidth}
        \centering
        \includegraphics[width=\linewidth]{
        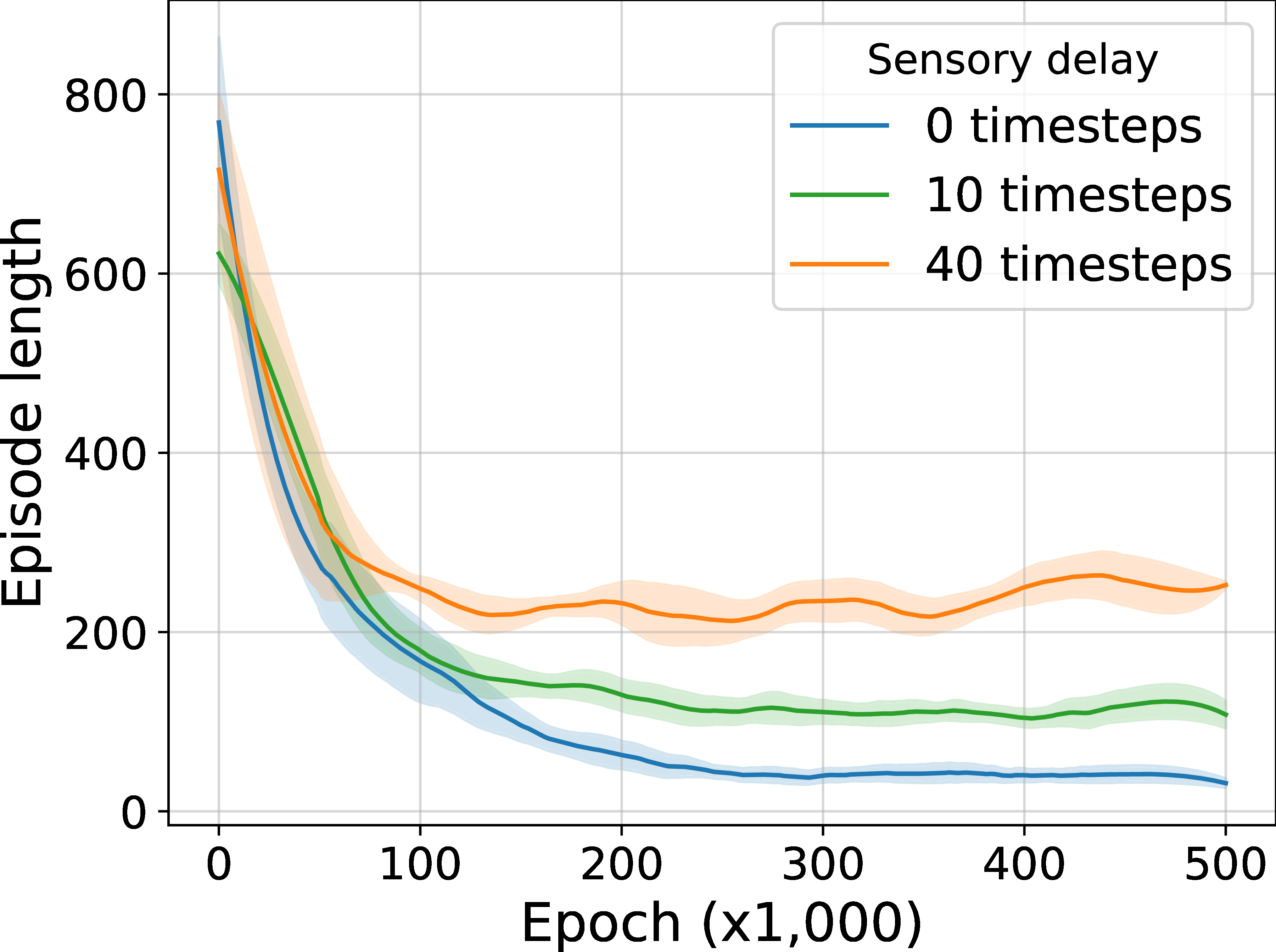
        }
        \caption{Time to reach target vs epoch}
    \end{subfigure}
    \hfill
    \begin{subfigure}[t]{0.21\textwidth}
        \centering
        \includegraphics[width=\linewidth]{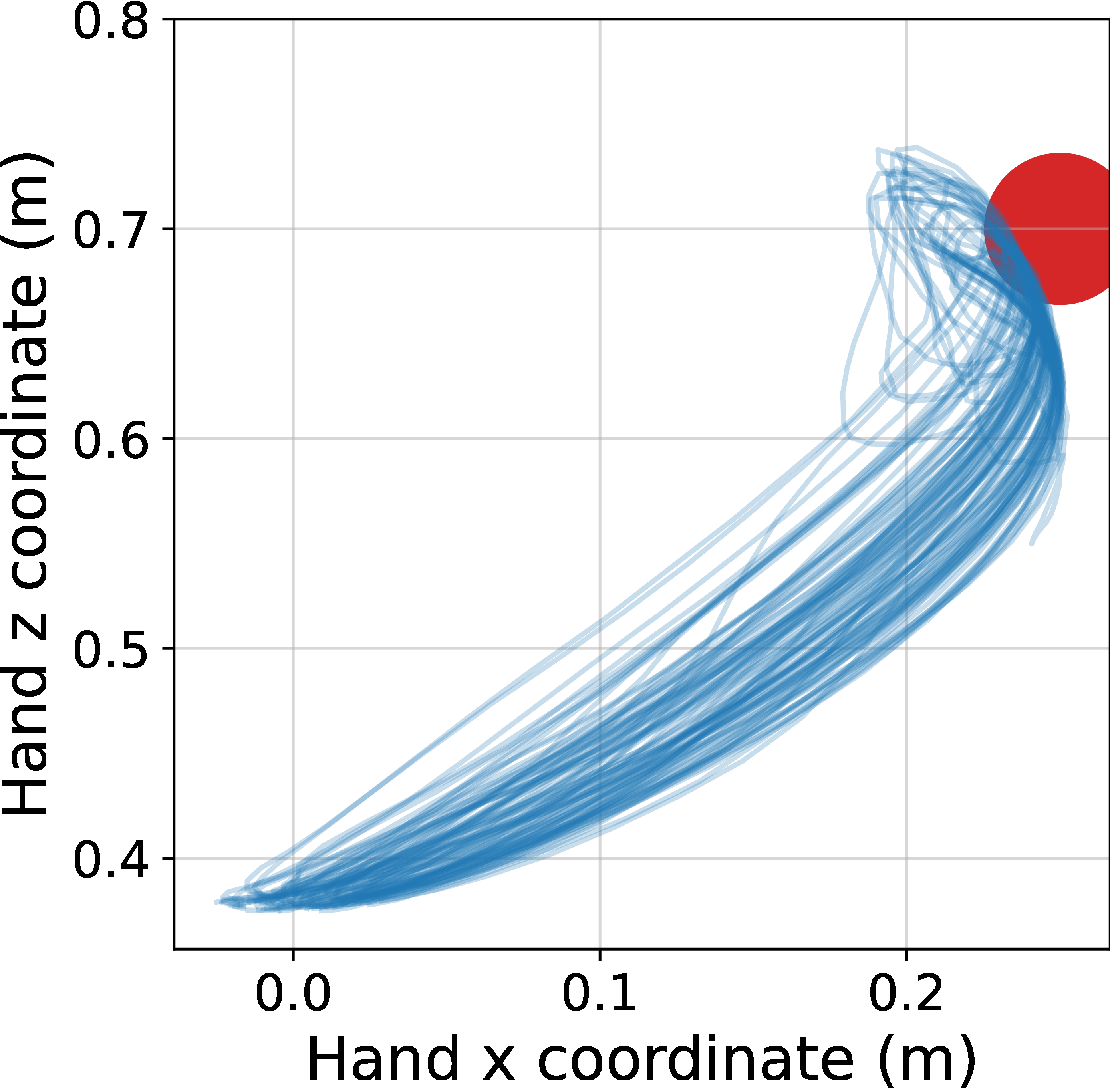}
        \caption{Trajectories without delay}
    \end{subfigure}
    \hfill
    \begin{subfigure}[t]{0.21\textwidth}
        \centering
        \includegraphics[width=\linewidth]{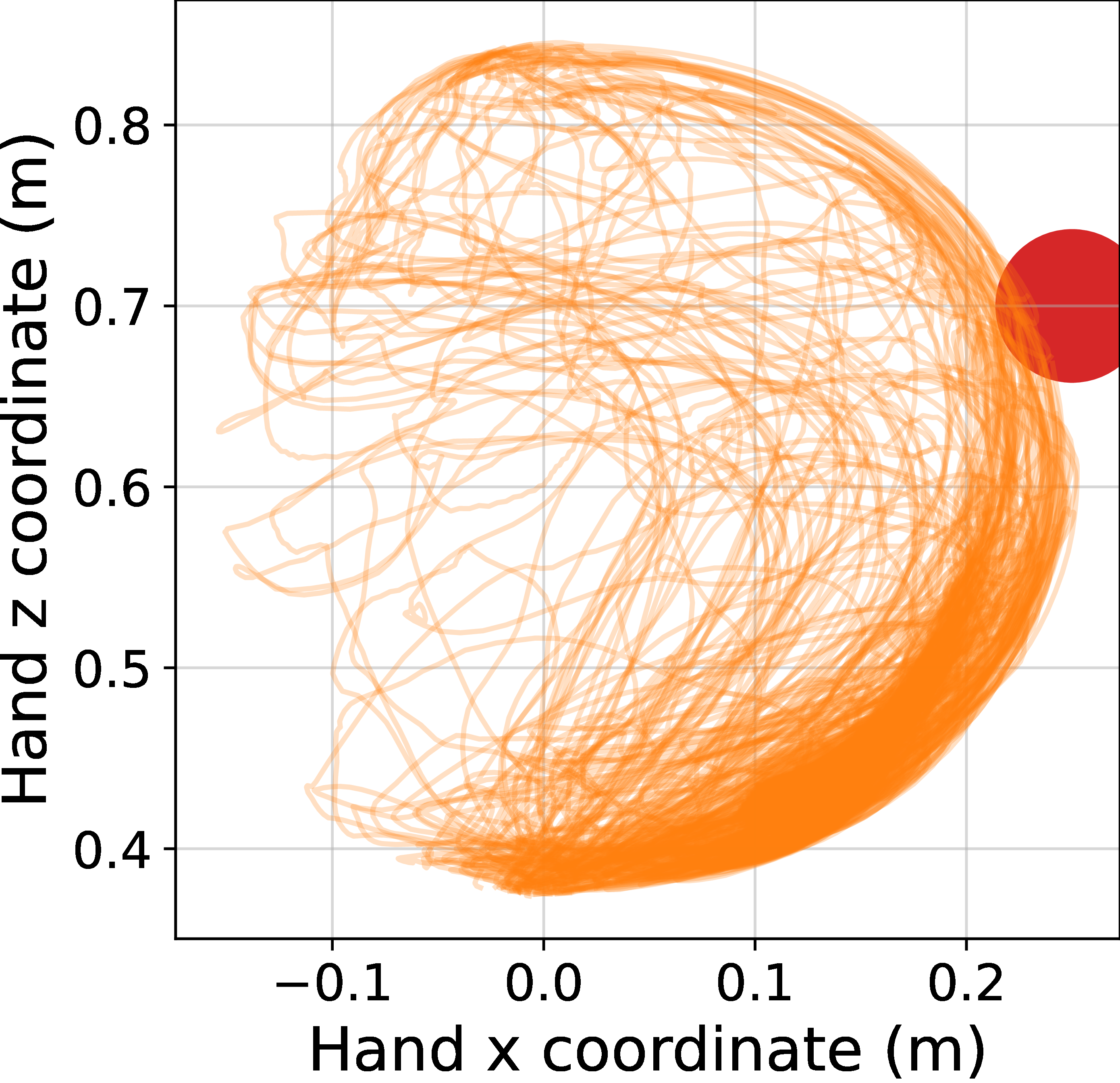}
        \caption{Trajectories with delay of 40 timesteps, i.e. 200~ms}
    \end{subfigure}
    \hfill
    \caption{Sensorimotor delay experiments. (a) Illustration of the reaching task. (b) Learning curves for different sensorimotor delays. (c) Example hand trajectories after learning without delays. (d) Same as (c) but for a delay of 200 ms.}
    \label{fig:delays}
\end{figure*}

\section{Extended Sensory Modalities}

\subsection{Developing visual acuity}

Babies are born with immature vision, which improves over the first few months of life. In particular they have low sensitivity to high spatial frequencies, resulting in blurry vision. This is characterized by the Contrast Sensitivity Function (CSF) \cite{kiorpes2016puzzle}. Relative to adults, infants have a lower CSF for the entire spectrum and their sensitivity to spatial frequencies higher than a cutoff value known as the visual acuity is negligible.

Since MIMo uses MuJoCo's default renderer, which is optimized for performance rather than realism, we opt for a simple model of the developing visual acuity. First, we interpolate the monocular visual acuities reported by Neijzen et al.\ for infants between 1 and 24 months \cite{neijzen2025reference}, so as to determine the acuity for a MIMo of any age. Then, we approximate the CSF as linear functions given by:
\begin{equation} \label{eq:csf}
    CSF_a (f) = \left\{
        \begin{array}{ c l }
        1 - f / a & \textrm{if\, \(f < a\),} \\
        0 & \textrm{otherwise}
    \end{array}
    \right. \; ,
\end{equation}

\noindent where \(f\) is the spatial frequency measured in cycles per degree and the visual acuity \(a\) depends on  MIMo's age (see Fig.~\ref{fig:acuity}(d). This function acts as a smooth low-pass filter on the visual observations: to apply it, the images are converted to the frequency domain with a fast Fourier transform, where each value in the frequency space is multiplied by the corresponding CSF, and then the image is converted back to pixel space. 

While this implementation of the CSF is simple and does not account for low sensitivity in the range of low spatial frequencies, it permits an efficient way to explore the consequences of infants' developing contrast sensitivity (see Fig.~\ref{fig:acuity}(e)--(i)). 

\subsection{Foveation}

Another important aspect of human vision is foveation, i.e. the higher density of photoreceptors in the center of the retina compared to the periphery \cite{wells2016variation}, with its corresponding cortical magnification, i.e. larger areas of primary visual cortex used to process the central region of the field of view \cite{duncan2003cortical}. Foveation can account for eccentricity-dependent acuity \cite{duncan2003cortical} mechanisms and eye movements \cite{raabe2023saccade, lopez2024self}.

A common approach for modeling foveation is to render multiple concentric views of increasing field-of-view at decreasing resolutions, as implemented for MIMo in \cite{raabe2023saccade}. However, this comes at an increased computational cost. Here, we implement an alternative solution: MIMo's foveated vision is achieved by rendering at a single resolution and field-of-view in cartesian space, converting to a log-polar space, and inverting the polar mapping back to cartesian space while retaining the logarithmic dependency of the eccentricity. As seen in Fig.~\ref{fig:acuity}(c), this expands the visual area assigned to the central region and compresses the periphery, in line with the human retina \cite{wells2016variation} and primary visual cortex \cite{duncan2003cortical}. 

\subsection{Sensorimotor delays}

Signals from the sensory periphery require finite time to travel to sensory processing areas in the brain. Similarly, motor commands generated in the brain take time to travel to the muscles.
Such delays contribute to humans' finite reaction times.
Estimates consider minimum reaction times of between 50 and 100~milliseconds for proprioception-driven actions \cite{manning2012proprioceptive} and larger times for exteroceptive sensory modalities, and although these vary between sensory modalities the fastest reaction times are around 200~milliseconds in adults \cite{whelan2008effective}.

Here we introduce a simple but efficient way of modeling such delays. Specifically, we introduce first-in-first-out buffers for sensory signals from all modalities and for motor commands. The present observation measured from MIMo's sensory modules is stored, and the observation that was collected a given number of timesteps ago is returned. Likewise, the present motor command selected by MIMo is stored, and a motor command selected previously is executed. This allows us to model the different conduction velocity maturation times of the sensory and the motor neural fibers \cite{garcia2004neurophysiology}. By default MIMo simulations are performed with a timestep of 5~ms, and the sensorimotor delays defined by the user are taken in multiples of this timestep duration. This distinction 

To investigate the consequences of such sensorimotor delays for learning, we recreate the reaching experiments presented in \cite{mattern2024mimo}, where MIMo needs to reach a target presented in front of him with his right hand (see Fig.~\ref{fig:delays}). MIMo is trained with the Proximal-Policy Optimization (PPO) reinforcement learning algorithm, trying to maximize an extrinsic reward for touching the target, for a total of 500,000 epochs. We compare 3 alternative sensory delays: no delay, a delay of 50~ms, and a delay of 200~ms. Since the environment is static, i.e. the target does not move, motor delays are set to the minimum of one timestep or 5~ms. The results reveal that MIMo is able to achieve the task in all cases, but it takes him longer to do so with higher delays. Furthermore, an analysis of the resulting hand trajectories reveals that the introduction of delays results in noisy movements somewhat reminiscent of {\em ataxia}. Humans are thought to compensate for such delays by using forward models involving the cerebellum that we have omitted here.

\begin{figure}[!t]
    \centering
    \begin{subfigure}[t]{0.45\linewidth}
        \centering
        \includegraphics[width=\linewidth]{
        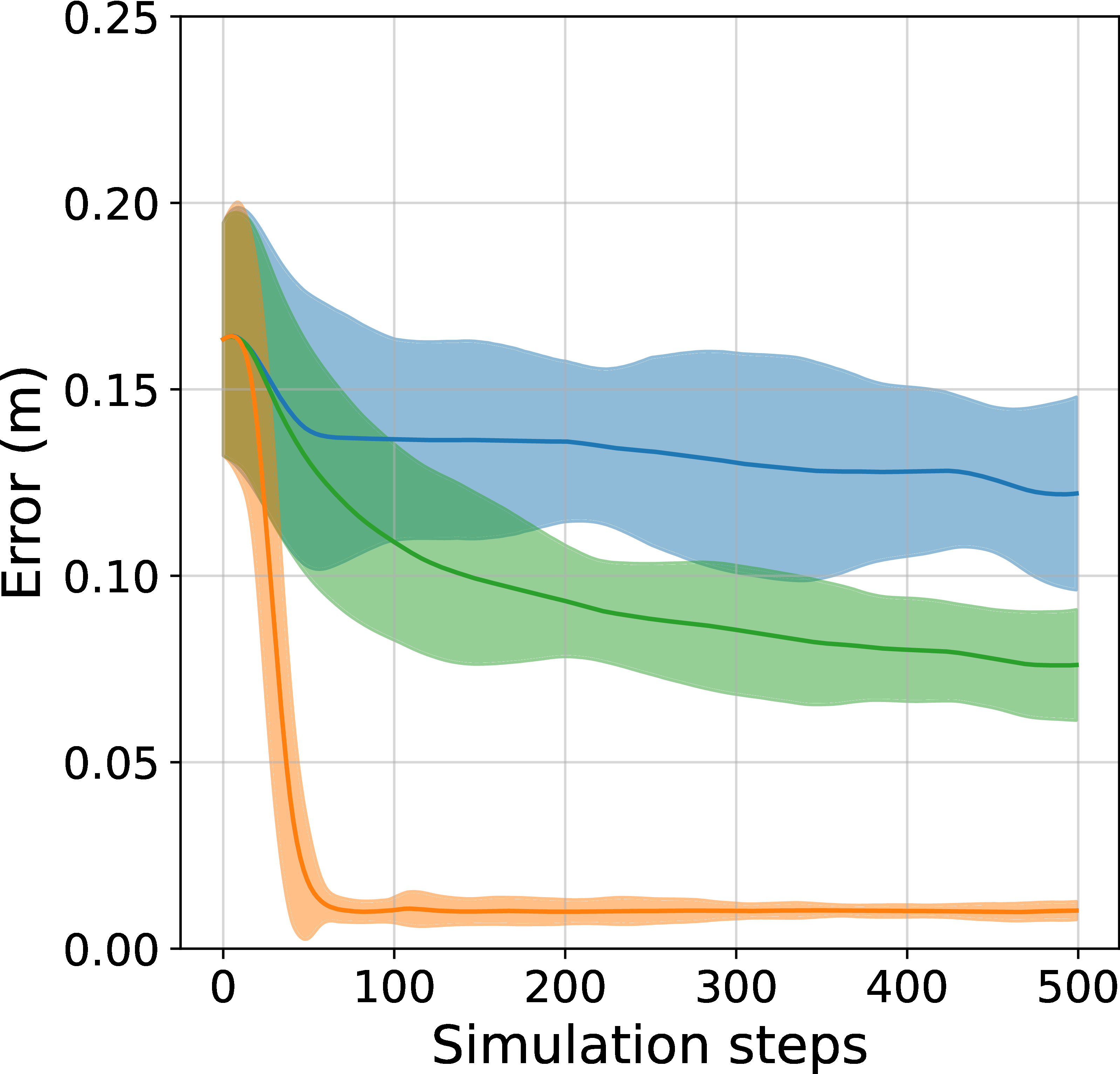
        }
        \caption{Hand-target distance}
    \end{subfigure}
    \hfill
    \begin{subfigure}[t]{0.45\linewidth}
        \centering
        \includegraphics[width=\linewidth]{
        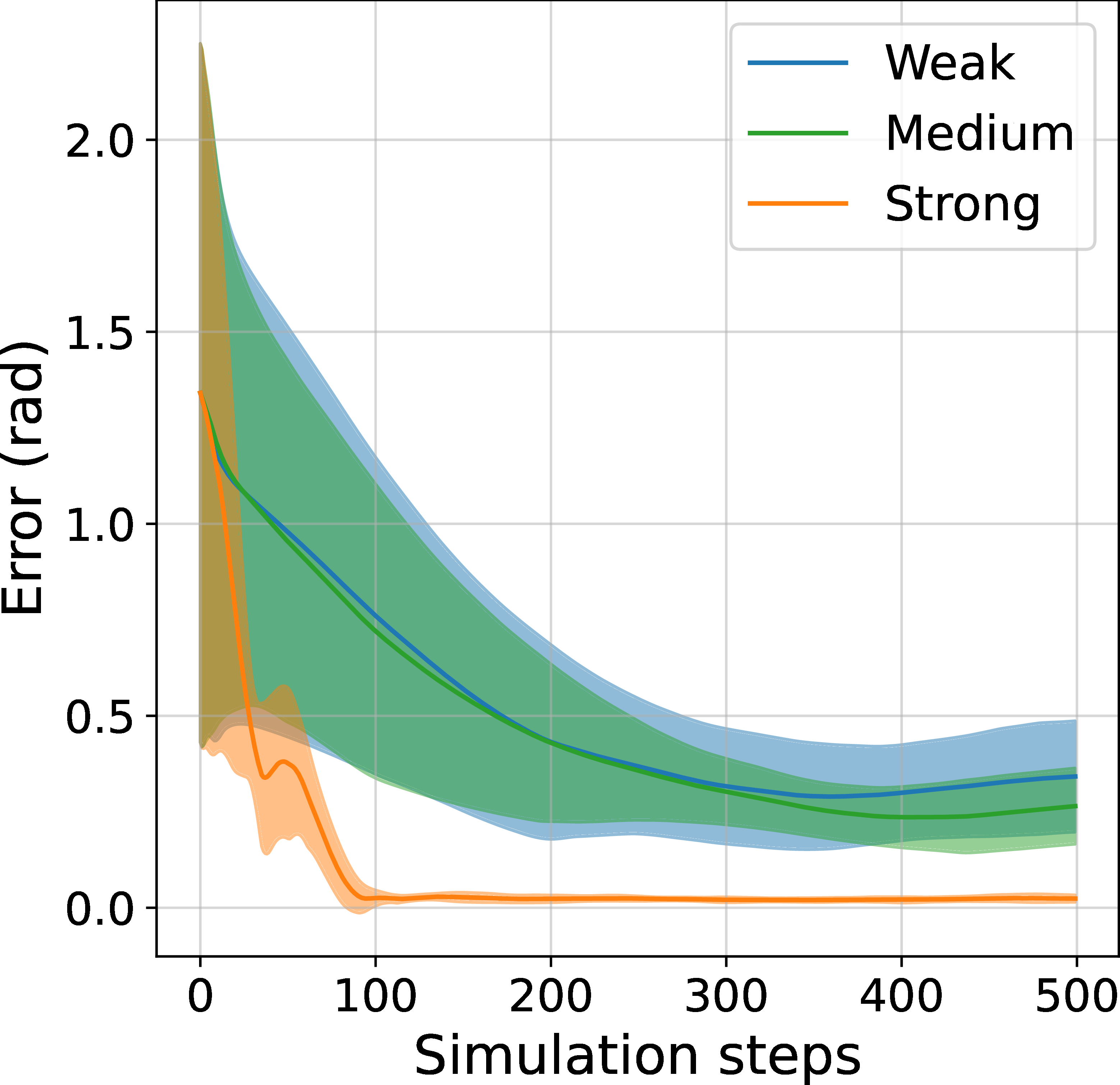
        }
        \caption{Eye-target misalignment}
    \end{subfigure}
    
    \caption{Example of the inverse dynamics controller where MIMo's hand (eyes) reaches towards (align with) a target. Errors are averaged across 10 random reachable locations for each of three increasingly-stronger sets of attractive forces.}
    \label{fig:inverse}
\end{figure}

\section{Other Updates}

\subsection{Inverse dynamics control}

MIMo is primarily designed as a simulation platform for learning experiments, and as such he is expected to learn his movement patterns by exploiting his actuation models. However, there might be instances in which it is necessary to precisely control MIMo's body, e.g. to retarget a real infant's motion or to gather demonstrations for imitation learning. 
MuJoCo provides a simple way to control body parts via \textit{mocap bodies}—virtual markers that pull parts of the agent toward target locations. However, this method affects all joints involved in the movement, making it difficult to precisely control only specific parts of the body while keeping others fixed. 

To address this, we develop an inverse dynamics controller for MIMo v2, allowing targeted control over individual joints or groups of joints. Instead of simply dragging an end effector toward a position, this controller computes the forces required to exclusively move the intended joints in order to reach the desired pose in operational space.
The desired behavior is defined via a prioritized sequence of tasks (e.g. reach or align to a point in space). The task priority algorithm ensures that a task will not interfere with those at higher priority, a necessary condition for safety related tasks such as obstacle avoidance \cite{nakamura1987task}.
The controller formulation is an extension of the one presented in \cite{mistry2012operational} for underactuated systems, itself a streamlined representation of the original one by Khatib \cite{khatib1987unified} (see MIMo's documentation for details). End effector sites can be inserted within any point of the kinematic chain specified in MuJoCo, such as MIMo's hands or eyes. Each of these is linked to a virtual target in operational space and guided by a polynomial error function given their relative position and rotation. The error metric can be any of the following: translational distance, to move towards a point in space; orientation error, to match a reference orientation; or vector misalignment, to align to the distance vector between the end effector site and the target. Fig.~\ref{fig:inverse} shows an example usage of the inverse dynamics controller, in which the MIMo's hands have to reach a random target in front of MIMo, while the eyes align with it.

\subsection{Procedural environment generator}

\begin{figure}[!b]
    \centering
    \includegraphics[width=.23\textwidth]{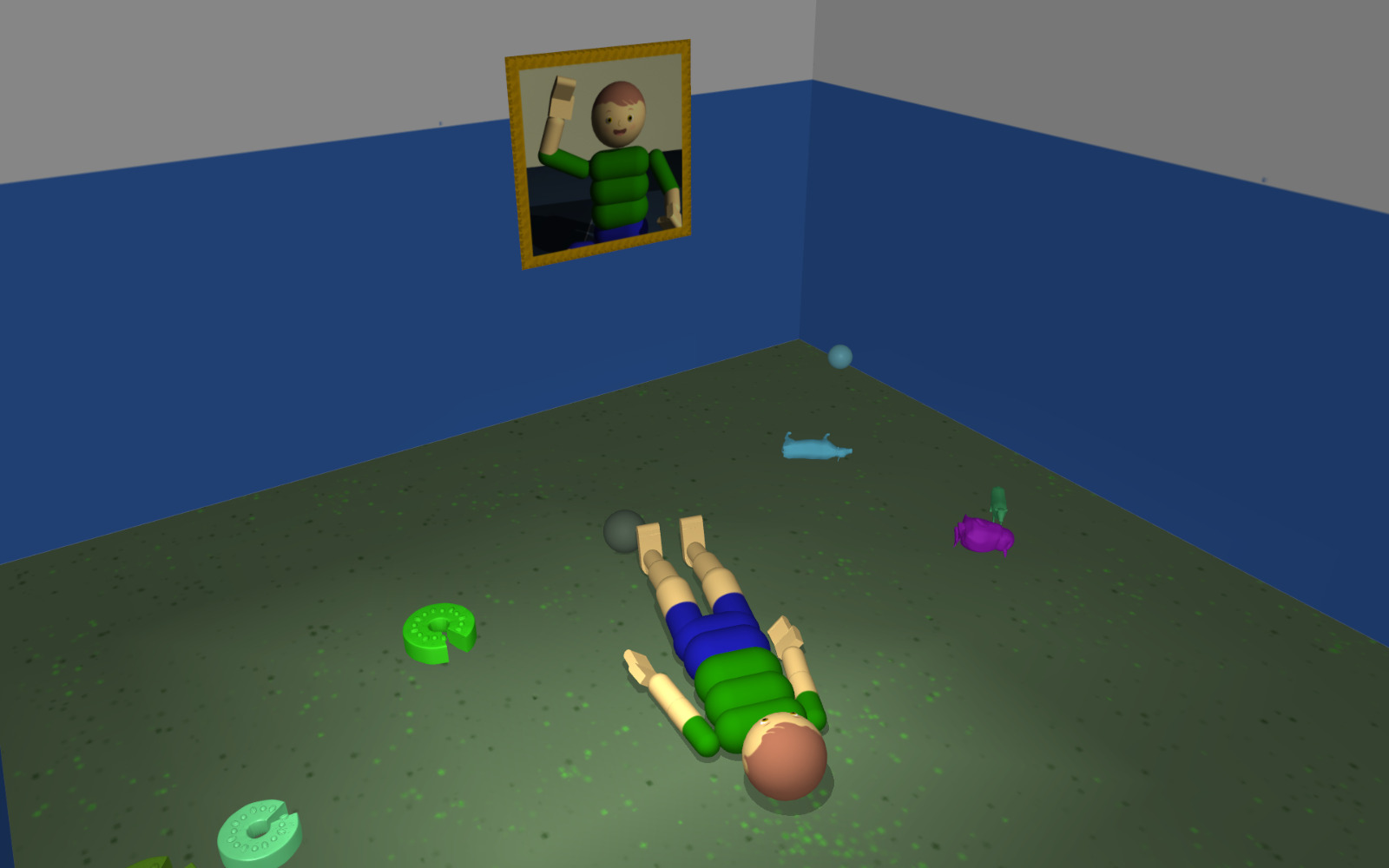}
    \hfill
    \includegraphics[width=.23\textwidth]{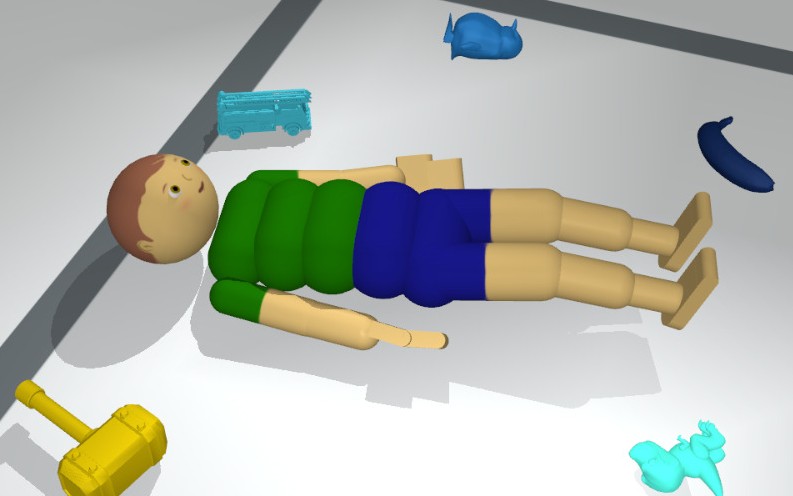}
    \caption{Examples of MIMo's random environment generator with various room textures and objects from Toys4K \cite{stojanov21cvpr}. }
    \label{fig:environment}
\end{figure}

A key challenge in learning various skills is the generalization of behaviors to novel contexts. This can benefit from training in a variety of environments. To support this, we develop a scene randomizer for MIMo v2, generating rooms of varying sizes, with floor, ceiling, and wall textures randomly sampled from an expandable list.
A room is comprised of 6 planes (floor, ceiling, walls) randomly sized. Their textures are individually sampled from a list that can be expanded arbitrarily. MIMo's location is then chosen randomly to lie within the room borders.
A random number of toys is sampled from a subset of the Toys4K dataset \cite{stojanov21cvpr}, and a random color is applied to each. The toys are distributed around MIMo, following a distribution that encourages locations close to reachable areas.
Reproducibility is guaranteed by controlling the seed of the random generation process. Examples of the random environments are shown in Fig.~\ref{fig:environment}.

\subsection{Updated compatibility}

MIMo v2 is updated to benefit from recent upgrades of MuJoCo v3 and Gymnasium, which enables faster simulation. 
Environment generation is now modular, allowing users to combine different sensory modalities, actuation models, and scenes.
Additionally, in order to make it easier to use MIMo out-of-the-box in different operating systems and allow for guaranteed reproducibility of experiments, we provide Singularity containers where all the necessary third-party libraries are provided without requiring installation.

\section{Conclusions and Perspectives}

We have presented a major update to the MIMo simulation platform including physical growth, the development of visual acuity, foveated vision, and sensorimotor delays. This is the first virtual embodiment that allows researchers to study the consequences of bodily and sensory changes during infancy. In ongoing work, MIMo v2 is being used to model complex behaviors, including crawling and self-touch. A number of future extensions of the MIMo platform are still desirable, in particular the incorporation of additional sensory modalities such as olfaction, audition, and nociception (the latter two already being work in progress). Furthermore, MIMo should be able to interact with social partners to study aspects of social and communicative development. 

Clearly, a simulation platform such as MIMo only allows for an approximation of the physical and physiological details of an infant interacting with their environment: the task of creating a true {\em digital twin} of a developing infant is simply too daunting. Nevertheless, we believe that MIMo v2 (and even more so future ones) will be able to capture the {\em essence} of such interactions in sufficient detail. As we have argued before \cite{mattern2024mimo}, decisive advantages of such a simulation approach are the ease of use, low cost, high level of reproducibility, and facilitation of cumulative model construction compared to (developmental) robotics approaches. Despite its limitations, we hope that MIMo v2 will be a valuable tool for modeling and understanding the processes underlying early cognitive development and the emergence of consciousness in human infants.

\addtolength{\textheight}{-12cm}   








\begin{thebibliography}{10}
\providecommand{\url}[1]{#1}
\csname url@samestyle\endcsname
\providecommand{\newblock}{\relax}
\providecommand{\bibinfo}[2]{#2}
\providecommand{\BIBentrySTDinterwordspacing}{\spaceskip=0pt\relax}
\providecommand{\BIBentryALTinterwordstretchfactor}{4}
\providecommand{\BIBentryALTinterwordspacing}{\spaceskip=\fontdimen2\font plus
\BIBentryALTinterwordstretchfactor\fontdimen3\font minus
  \fontdimen4\font\relax}
\providecommand{\BIBforeignlanguage}[2]{{%
\expandafter\ifx\csname l@#1\endcsname\relax
\typeout{** WARNING: IEEEtran.bst: No hyphenation pattern has been}%
\typeout{** loaded for the language `#1'. Using the pattern for}%
\typeout{** the default language instead.}%
\else
\language=\csname l@#1\endcsname
\fi
#2}}
\providecommand{\BIBdecl}{\relax}
\BIBdecl

\bibitem{winstanley2023stages}
M.~A. Winstanley, ``Stages in theory and experiment. fuzzy-structuralism and
  piagetian stages,'' \emph{Integrative Psychological and Behavioral Science},
  vol.~57, no.~1, pp. 151--173, 2023.

\bibitem{anthrokids}
S.~Ressler, ``{A}nthro{K}ids - {A}nthropometric data of children,'' \emph{Nat.
  Inst. Standards and Technol.}, 1977.

\bibitem{corbetta2021perception}
D.~Corbetta, ``Perception, action, and intrinsic motivation in infants’
  motor-skill development,'' \emph{Current Directions in Psychological
  Science}, vol.~30, no.~5, pp. 418--424, 2021.

\bibitem{tondel2008accommodation}
G.~M. Tondel and T.~R. Candy, ``Accommodation and vergence latencies in human
  infants,'' \emph{Vision research}, vol.~48, no.~4, pp. 564--576, 2008.

\bibitem{corbetta2018learning}
D.~Corbetta, R.~F. Wiener, and S.~L. Thurman, ``Learning to reach in infancy,''
  \emph{Reach-to-Grasp behavior}, pp. 18--41, 2018.

\bibitem{harbourne2013sit}
R.~T. Harbourne, M.~A. Lobo, G.~M. Karst, and J.~C. Galloway, ``Sit happens:
  Does sitting development perturb reaching development, or vice versa?''
  \emph{Infant Behavior and Development}, vol.~36, no.~3, pp. 438--450, 2013.

\bibitem{adolph2017development}
K.~E. Adolph and J.~M. Franchak, ``The development of motor behavior,''
  \emph{Wiley Interdisciplinary Reviews: Cognitive Science}, vol.~8, no. 1-2,
  p. e1430, 2017.

\bibitem{metta2010icub}
G.~Metta, L.~Natale, F.~Nori, G.~Sandini, D.~Vernon, L.~Fadiga, C.~Von~Hofsten,
  K.~Rosander, M.~Lopes, J.~Santos-Victor \emph{et~al.}, ``The icub humanoid
  robot: An open-systems platform for research in cognitive development,''
  \emph{Neural networks}, vol.~23, no. 8-9, pp. 1125--1134, 2010.

\bibitem{kerzel2017nico}
M.~Kerzel, E.~Strahl, S.~Magg, N.~Navarro-Guerrero, S.~Heinrich, and
  S.~Wermter, ``Nico—neuro-inspired companion: A developmental humanoid robot
  platform for multimodal interaction,'' in \emph{2017 26th IEEE International
  Symposium on Robot and Human Interactive Communication (RO-MAN)}.\hskip 1em
  plus 0.5em minus 0.4em\relax IEEE, 2017, pp. 113--120.

\bibitem{mattern2024mimo}
D.~Mattern, P.~Schumacher, F.~M. L{\'o}pez, M.~C. Raabe, M.~R. Ernst,
  A.~Aubret, and J.~Triesch, ``Mimo: A multimodal infant model for studying
  cognitive development,'' \emph{IEEE Transactions on Cognitive and
  Developmental Systems}, vol.~16, no.~4, pp. 1291--1301, 2024.

\bibitem{kim2022simulating}
D.~Kim, H.~Kanazawa, and Y.~Kuniyoshi, ``Simulating a human fetus in soft
  uterus,'' in \emph{2022 IEEE International Conference on Development and
  Learning (ICDL)}.\hskip 1em plus 0.5em minus 0.4em\relax IEEE, 2022, pp.
  135--141.

\bibitem{worldhealthorganization2006WHO}
{World Health Organization}, ``{WHO} child growth standards:
  Length/height-for-age, weight-for-age, weight-for-length, weight-for-height
  and body mass index-for-age: Methods and development,'' World Health
  Organization, Tech. Rep., Nov. 2006.

\bibitem{worldhealthorganization2007WHO}
------, ``{WHO} child growth standards : Head circumference-for-age, arm
  circumference-for-age, triceps skinfold-for-age and subscapular
  skinfold-for-age : Methods and development,'' World Health Organization,
  Tech. Rep., Nov. 2007.

\bibitem{lopez2024self}
F.~M. L{\'o}pez, M.~C. Raabe, B.~E. Shi, and J.~Triesch, ``Self-calibrating
  saccade-vergence interactions,'' in \emph{2024 IEEE International Conference
  on Development and Learning (ICDL)}.\hskip 1em plus 0.5em minus 0.4em\relax
  IEEE, 2024, pp. 1--7.

\bibitem{raabe2023saccade}
M.~C. Raabe, F.~M. L{\'o}pez, Z.~Yu, S.~Caplan, C.~Yu, B.~E. Shi, and
  J.~Triesch, ``Saccade amplitude statistics are explained by cortical
  magnification,'' in \emph{2023 IEEE International Conference on Development
  and Learning (ICDL)}.\hskip 1em plus 0.5em minus 0.4em\relax IEEE, 2023, pp.
  300--305.

\bibitem{fiala2023retargeting}
O.~Fiala, ``Retargeting infant movements to baby humanoid robots,'' 2023.

\bibitem{lopez2023eye}
F.~M. L{\'o}pez, B.~E. Shi, and J.~Triesch, ``Eye-hand coordination develops
  from active multimodal compression,'' in \emph{2023 IEEE International
  Conference on Development and Learning (ICDL)}.\hskip 1em plus 0.5em minus
  0.4em\relax IEEE, 2023, pp. 437--442.

\bibitem{todorov2012mujoco}
E.~Todorov, T.~Erez, and Y.~Tassa, ``Mujoco: A physics engine for model-based
  control,'' in \emph{2012 IEEE/RSJ international conference on intelligent
  robots and systems}.\hskip 1em plus 0.5em minus 0.4em\relax IEEE, 2012, pp.
  5026--5033.

\bibitem{towers2024gymnasium}
M.~Towers, A.~Kwiatkowski, J.~Terry, J.~U. Balis, G.~De~Cola, T.~Deleu,
  M.~Goulao, A.~Kallinteris, M.~Krimmel, A.~KG \emph{et~al.}, ``Gymnasium: A
  standard interface for reinforcement learning environments,'' \emph{arXiv
  preprint arXiv:2407.17032}, 2024.

\bibitem{mattern2022mimo}
D.~Mattern, F.~M. López, M.~R. Ernst, A.~Aubret, and J.~Triesch, ``{MIM}o: A
  multi-modal infant model for studying cognitive development in humans and
  {AI}s,'' 2022, pp. 23--29.

\bibitem{eliks2022alberta}
M.~Eliks and E.~Gajewska, ``The alberta infant motor scale: A tool for the
  assessment of motor aspects of neurodevelopment in infancy and early
  childhood,'' \emph{Frontiers in neurology}, vol.~13, p. 927502, 2022.

\bibitem{jaric2003role}
S.~Jaric, ``Role of body size in the relation between muscle strength and
  movement performance,'' \emph{Exercise and sport sciences reviews}, vol.~31,
  no.~1, pp. 8--12, 2003.

\bibitem{van2021active}
P.~A. van Iersel and M.~Hadders-Algra, ``Active head lifting from supine in
  infancy in the general population: Red flag or not?'' \emph{Early Human
  Development}, vol. 163, p. 105466, 2021.

\bibitem{angulo2002three}
R.~M. Angulo-Kinzler, B.~Ulrich, and E.~Thelen, ``Three-month-old infants can
  select specific leg motor solutions,'' \emph{Motor Control}, vol.~6, no.~1,
  pp. 52--68, 2002.

\bibitem{kiorpes2016puzzle}
L.~Kiorpes, ``The puzzle of visual development: behavior and neural limits,''
  \emph{Journal of Neuroscience}, vol.~36, no.~45, pp. 11\,384--11\,393, 2016.

\bibitem{neijzen2025reference}
C.~M. Neijzen, F.~M. de~Wit, Y.~M. Hettinga, J.~H. de~Boer, M.~M. van Genderen,
  and G.~C. de~Wit, ``Reference values for the teller acuity cards ii (tac ii)
  in infants and preverbal children, a meta-analysis,'' \emph{Acta
  Ophthalmologica}, 2025.

\bibitem{wells2016variation}
E.~Wells-Gray, S.~Choi, A.~Bries, and N.~Doble, ``Variation in rod and cone
  density from the fovea to the mid-periphery in healthy human retinas using
  adaptive optics scanning laser ophthalmoscopy,'' \emph{Eye}, vol.~30, no.~8,
  pp. 1135--1143, 2016.

\bibitem{duncan2003cortical}
R.~O. Duncan and G.~M. Boynton, ``Cortical magnification within human primary
  visual cortex correlates with acuity thresholds,'' \emph{Neuron}, vol.~38,
  no.~4, pp. 659--671, 2003.

\bibitem{manning2012proprioceptive}
C.~Manning, S.~Tolhurst, and P.~Bawa, ``Proprioceptive reaction times and
  long-latency reflexes in humans,'' \emph{Experimental brain research}, vol.
  221, no.~2, pp. 155--166, 2012.

\bibitem{whelan2008effective}
R.~Whelan, ``Effective analysis of reaction time data,'' \emph{The
  psychological record}, vol.~58, pp. 475--482, 2008.

\bibitem{garcia2004neurophysiology}
A.~Garc{\'\i}a-Garc{\'\i}a and J.~Calleja-Fern{\'a}ndez, ``Neurophysiology of
  the development and maturation of the peripheral nervous system,''
  \emph{Revista De Neurologia}, vol.~38, no.~1, pp. 79--83, 2004.

\bibitem{nakamura1987task}
Y.~Nakamura, H.~Hanafusa, and T.~Yoshikawa, ``Task-priority based redundancy
  control of robot manipulators,'' \emph{The International Journal of Robotics
  Research}, vol.~6, no.~2, pp. 3--15, 1987.

\bibitem{mistry2012operational}
M.~Mistry and L.~Righetti, ``Operational space control of constrained and
  underactuated systems,'' in \emph{Robotics: Science and systems}, vol.~7,
  2012, pp. 225--232.

\bibitem{khatib1987unified}
O.~Khatib, ``A unified approach for motion and force control of robot
  manipulators: The operational space formulation,'' \emph{IEEE Journal on
  Robotics and Automation}, vol.~3, no.~1, pp. 43--53, 1987.

\bibitem{stojanov21cvpr}
S.~Stojanov, A.~Thai, and J.~M. Rehg, ``Using shape to categorize: Low-shot
  learning with an explicit shape bias,'' 2021.

\end{thebibliography}
\end{document}